\documentclass[10pt,twocolumn,letterpaper]{article}

\usepackage{mathtools}

\DeclarePairedDelimiter\floor{\lfloor}{\rfloor}
\usepackage{cvpr}
\usepackage{times}
\usepackage{epsfig}
\usepackage{graphicx}
\usepackage{amsmath}
\usepackage{amssymb}
\usepackage{graphicx}

\usepackage[breaklinks=true,bookmarks=false]{hyperref}

\cvprfinalcopy 

\def\cvprPaperID{****} 

\begin{document}

\title{Improving Context-Aware Semantic Relationships in Sparse Mobile Datasets}

\author{Peter Hansel\\
Stanford University\\
{\tt\small pwhansel@stanford.edu}
\and Nik Marda\\
Stanford University\\
{\tt\small nmarda@stanford.edu}
\and William Yin\\
Stanford University\\
{\tt\small wyin@stanford.edu}
}

\maketitle

\begin{abstract}
Traditional semantic similarity models often fail to encapsulate the external context in which texts are situated. However, textual datasets generated on mobile platforms can help us build a truer representation of semantic similarity by introducing multimodal data. This is especially important in sparse datasets, making solely text-driven interpretation of context more difficult. In this paper, we develop new algorithms for building external features into sentence embeddings and semantic similarity scores. Then, we test them on embedding spaces on data from Twitter, using each tweet's time and geolocation to better understand its context. Ultimately, we show that applying PCA with eight components to the embedding space and appending multimodal features yields the best outcomes. This yields a considerable improvement over pure text-based approaches for discovering similar tweets. Our results suggest that our new algorithm can help improve semantic understanding in various settings.

\end{abstract}

\section{Introduction} 
Determining the semantic similarity between texts is an important task in practical NLP. New methods like Doc2Vec \cite{Doc2Vec} and Contextual Salience \cite{zel} achieve better results by incorporating context in computing semantic similarity. However, these methods still rely on solely textual features. When a dataset is sparse, these methods will perform even worse, as context is even harder to effectively determine.

However, with the advent of mobile devices, we often have access to a wealth of passively-collected information tied to any given piece of text, such as the time and location at which the text was recorded \cite{multimodal}\cite{yuan}. This multimodal data might provide valuable insight into the context of text input that cannot be captured by textual analysis alone. As a result, we might be able to develop more effective methods for determining semantic similarity. 

In this paper,
we demonstrate that these additional features can capture information about the context of a text, helping discover semantic similarities that traditional state-of-the-art methods often miss. We use data from popular microblogging website Twitter, which has temporal and geospatial features alongside short paragraphs of text, to test our method and demonstrate its effectiveness. The inputs to our algorithms are pairs of tweets along with their corresponding temporal and geospatial information. We then used iterative minimization and PCA-based techniques to output predicted semantic similarity scores for each tweet pair. These scores were compared against data labeled by hand by political science students.

\section{Related Work}

We build off of previous work that incorporates the context of a sentence to better determine semantic similarity between sentences. Staple models like Term Frequency-Inverse Document Frequency (tf-idf) struggle to incorporate contextual information. Instead, we let two more recent methods, Doc2Vec and Contextual Salience (CoSal), guide our approach toward building better contextual understanding.

Doc2Vec is a method that learns continuous distributed vector representations for inputted text, allowing it to better incorporate text ordering and semantics \cite{Doc2Vec}. This allows it to take the context of a document into account when computing similarity.


CoSal computes the importance of a word given its context. This is then used to produce weighted bag-of-words sentence embeddings, thereby incorporating context into semantic similarity computations. These contexts can also be small, as CoSal works well with as few as 20 sentences.

\section{Sample Demonstration on Twitter Data}
We tested our algorithms on the data.world Politician tweets dataset \cite{dataset}, a collection of over 1.5 million tweets from American federal politicians. We chose political tweets because we expected there to be meaningful temporal and geospatial information that text-based models would miss. Four Stanford political undergraduate students each manually labeled the semantic similarity between 360 pairs of randomly selected tweets from each of our models, assigning scores based on criteria of topical, ideological, and stylistic similarity. Then these scores were averaged and scaled to produce similarity labels between 0 and 1.

We preprocessed the data prior to constructing an embedding space. We began by associating each tweet with its corresponding user location. Then, using the GeoPy Nominatim API, we associated each location with corresponding longitude and latitude values. Next, we encoded time as cyclical continuous features\cite{website}, separating the timespans of one day from that of one year, and then maintaining the linear continuous feature form of multiple years. 
This made it possible to test our hypothesis that tweets that are closer during the cycle of a short period of time are more likely to be semantically similar. Finally, we stripped the tweets of URLs and stopwords and converted them all to lowercase. 

\section{Methods}
We tested two different approaches to including multimodal data in semantic similarity computations. Each of them build on related work described in Section 2.

\subsection{Iterative Minimization}

\subsubsection{Modifying Contextual Salience}
Our first approach was to directly improve upon the existing CoSal algorithm. We chose to work with CoSal because we believed it to be most suitable for sparse mobile datasets. The CoSal algorithm sends each sentence to a 50-dimensional embedding space using Mahalanobis distance over the context. It then computes the similarity score (not adjusting for context): 

$$sim_{CoSal}(a, b) = a \cdot b$$

In this method, our approach was to modify this equation to take into account additional features, such as geolocation and timestamp. In general, each input sentence $s$ can be represented by $n + 1$ features: $s = \{s_{CoSal}, s_1, s_2, ..., s_n\}$ where $s_{CoSal}$ is the vector encoding produced by the CoSal algorithm and each $s_i$ is an additional feature of the sentence (for example, in the Twitter data set $s_1$ is the time at which the Tweet was published and $s_2$ is an ordered pair representing longitude and latitude of where the Tweet was published). With these new inputs, we proposed two potential new functions as improvements over $sim_{CoSal}:$

\begin{equation}
\begin{split}
    \text{sim}_{\Sigma}(s^{(1)}, s^{(2)}) &= s^{(1)}_{CoSal} \cdot s^{(2)}_{CoSal} + \sum_{i=1}^n\alpha_id_i(s_i^{(1)}, s_i^{(2)})
\end{split}
\end{equation}

\begin{equation}
\begin{split}
    \text{sim}_{\Pi}(s^{(1)}, s^{(2)}) &= s^{(1)}_{CoSal} \cdot s^{(2)}_{CoSal}\left(\prod_{i=1}^n\alpha_i + d_i(s_i^{(1)}, s_i^{(2)})\right)
\end{split}
\end{equation}

\subsubsection{Defining Distance Formulas}

Each $d_i$ is a distance function with two basic properties: $\forall a,b$ we have $d_i(a,b) \in [0, 1]$ and when sentences are "closer" in a certain feature, their distance $d_i$ is smaller than the distance between two "further" sentences. Several candidate distance functions were tried for each $d_i$, such as:

\begin{equation}
d_i(a, b) = \exp{\left(-|a - b|\right)}
\end{equation}
and
\begin{equation}
d_i(a, b) = \frac{1}{|a - b| + 1}
\end{equation}

\subsubsection{Loss Function and Parameter Optimization}

Using these equations, a new similarity score was assigned to each pair of tweets. The model was trained and tested with batches of 10-20 tweets (smaller batches were necessary due to the difficulty of manually labeling all points).
The output of the model was an $m$ by $m$ matrix where each row corresponded to a tweet and each entry in the row was the ranking of similarity between that tweet and every other tweet. The loss function 

\begin{equation}
L(\alpha_1, \alpha_2) = \sqrt{\sum_{(s^{(i)}, s^{(j)})}{(\hat{y}(s^{(i)}, s^{(j)}) - y(s^{(i)}, s^{(j)}))^2}}
\end{equation}
calculated the difference between this matrix and the ranking matrix of the manually labeled data, where $y(a, b)$ is the ranking of a tweet from the labeled data and $\hat{y}(a, b)$ is the ranking from our function. This function was then minimized by varying $\alpha_1$ and $\alpha_2$. This was done by manual gradient descent: since the loss function is discrete, there is no well-defined gradient to use for traditional gradient descent.

\subsection{PCA and t-SNE}
\begin{figure}
    \centering
    \includegraphics[scale=0.6]{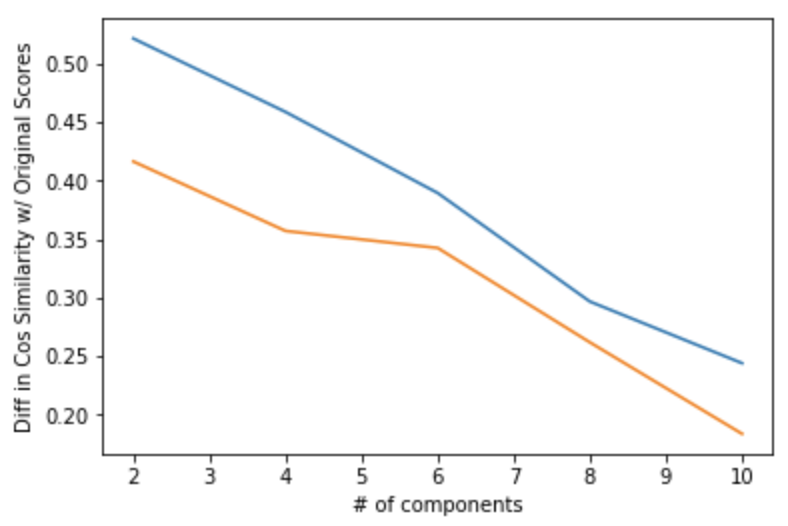}
    \caption{Average differences in cosine similarity score over 10 trials between the modified embedding space and the original embedding space. Blue is all appended features; orange is condensed time.}
    \label{cosinesim}
\end{figure}

We hypothesized that appending time and geolocation features to the Doc2Vec embedding space could induce closer semantic relationships. Recent work has suggested Doc2Vec works similarly to implicit matrix factorization \cite{neural}. Hence, we applied PCA as matrix factorization to the embedding space, artificially weighting the effect of the original space on the calculated similarity of word vectors. Note that the effect of the appended features is determined by the number of dimensions in the original embedding space, since the number of additional features is fixed. Figure \ref{cosinesim} shows that the difference in cosine similarity decreases as the number of components increases. 

We tested two ways of encoding time as a feature, shown in Figure \ref{bigboitimeboi}. The first included all appended features, encoding the two cyclical timescales of one day and one year, and the one linear timescale of multiple years. The second condensed time into a single value, representing all date/time values in terms of single values in seconds. In both variants, all features were standardized to zero mean and unit variance.

We also varied the number of components for the reduced tweet embedding space to determine which number of components produced the most realistic semantic similarity metric, according to similarity data labeled by Stanford political science students.  

Finally, we performed t-stochastic neighbor embedding (t-SNE) \cite{tsne} to reduce the embedding space to two dimensions for visualization. t-SNE is a method which first computes a joint probability distribution over each pair of vectors in the original space,

\begin{equation}
    p_{j|i} = \frac{\exp(-\|x_i - x_j\|^2/2\sigma_i^2)}{\sum_{k\neq i}\exp(-\|x_i-x_k\|^2/2\sigma_i^2)}
\end{equation}
as well as one over each pair of vectors in the two-dimensional space, 
\begin{equation}
    q_{j|i} = \frac{\exp{(-\|y_i-y_j\|^2)}}{\sum_{k\neq i}\exp(-\|y_i - y_k\|^2)}
\end{equation}
to then minimize the Kullback-Leibler divergences 
\begin{equation}
    C = \sum_i KL(P_i\|Q_i) = \sum_i\sum_j p_{j|i}\log\frac{p_{j|i}}{q_{j|i}}
\end{equation}
over all datapoints using gradient descent.

\section{Results and Discussion}

\subsection{Results of Iterative Minimization}

\subsubsection{Demonstrated Increase in Accuracy}

We performed gradient descent on our model, with many different combinations of function ($sim_\Sigma$ and $sim_\Pi$) with different $d_i$'s. We found that the loss function was minimized with the similarity function:

\begin{equation}
\begin{split}
	sim^*_\Pi(s^{(1)}, s^{(2)}) &= s^{(1)}_{CoSal} \cdot s^{(2)}_{CoSal}\\
    &\times (.02 + d_1(s^{(1)}_1, s^{(2)}_1)) \\
    &\times (9.55 + d_2(s^{(1)}_2, s^{(2)}_2))
\end{split}
\end{equation}

Where $d_1$ is the distance function $d_1(a, b) = \frac{1}{|a - b| + 1}$ where the inputs are the times (in units of days) when the tweets were published, and $d_2(a, b) = \frac{1}{10}(10 - \floor{\frac{\|a - b\|}{500}})$, where $\|a - b\|$ is the geographical distance between locations of tweets in miles.

Purely text-based CoSal similarity achieved an average loss of 32.80. Our best model achieved an average error of 29.86, for a loss decrease of 9.0 percent. In the next subsection, we discuss the inherent flaws of this model.

\begin{figure}
    \centering
    \includegraphics[scale=0.165]{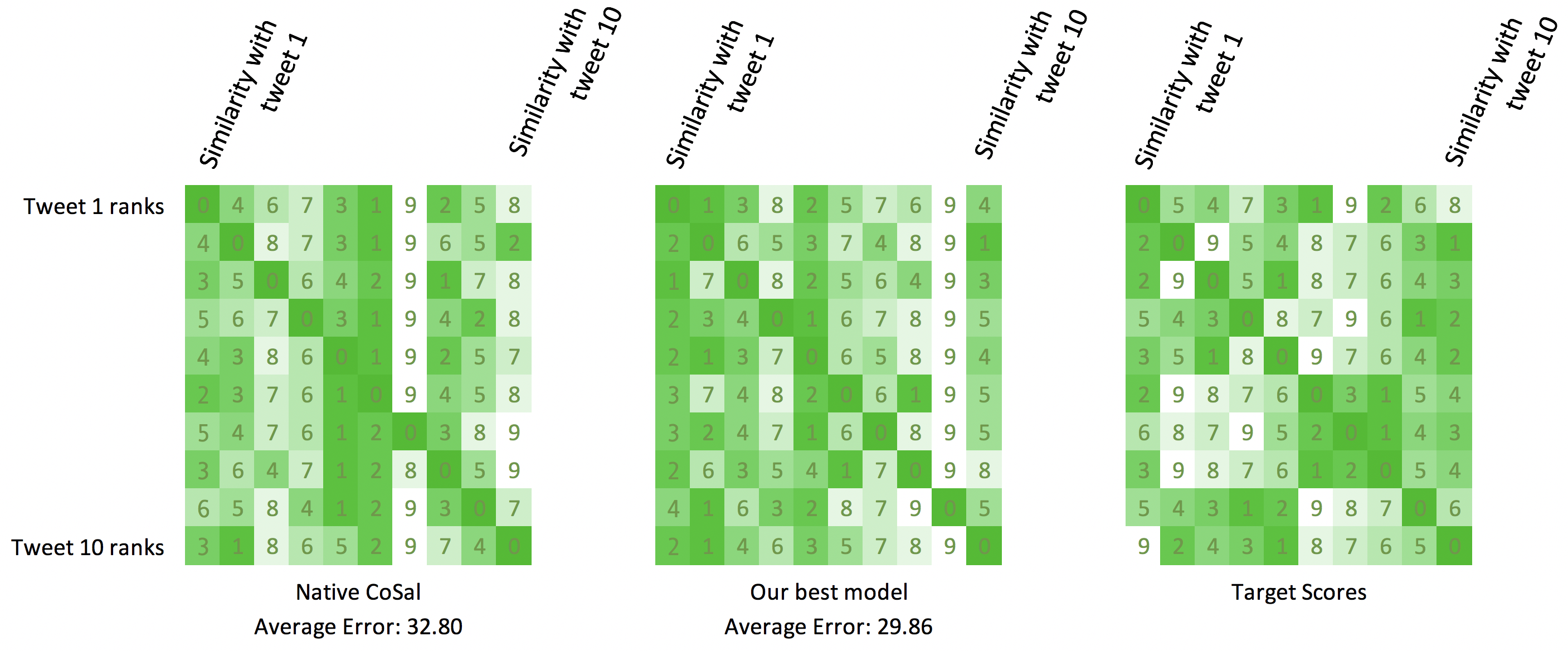}
    \caption{Comparing our algorithm against native CoSal and the labeled data. The $(i, j)$th entry in each grid is the number of tweets that are ranked more similar than tweet $j$ to tweet $i$.}
    \label{greengrid}
\end{figure}


\subsubsection{Limitations of this Approach}

Figure \ref{greengrid} displays the results CoSal and our best algorithm versus the labeled data rankings. We can observe that the rankings output by our model with this batch slightly better match the rankings from the labeled data. 
Notice that in the CoSal rankings matrix, many of the columns are almost entirely one color, meaning that corresponding Tweet was ranked nearly the same for most other tweets. This problem is slightly alleviated by incorporating other features (time and geolocation, in this case), but we see columns (such as column 9) that have largely uniform rankings. This is certainly not the case in the labeled data, and this problem is a trend in all data sets, regardless of exactly which functions we used in our optimization.
Ultimately, we can conclude that this method of modifying $sim_{CoSal}$ does indeed better predict similarity between sentences, but the level to which it can accurately do so is limited at a relatively low bar, possibly less than 10 percent better than unmodified CoSal.

\subsection{PCA and Visual Analysis}

\subsubsection{PCA Component Selection}
 
\begin{figure}
    \centering
    \includegraphics[scale=0.52]{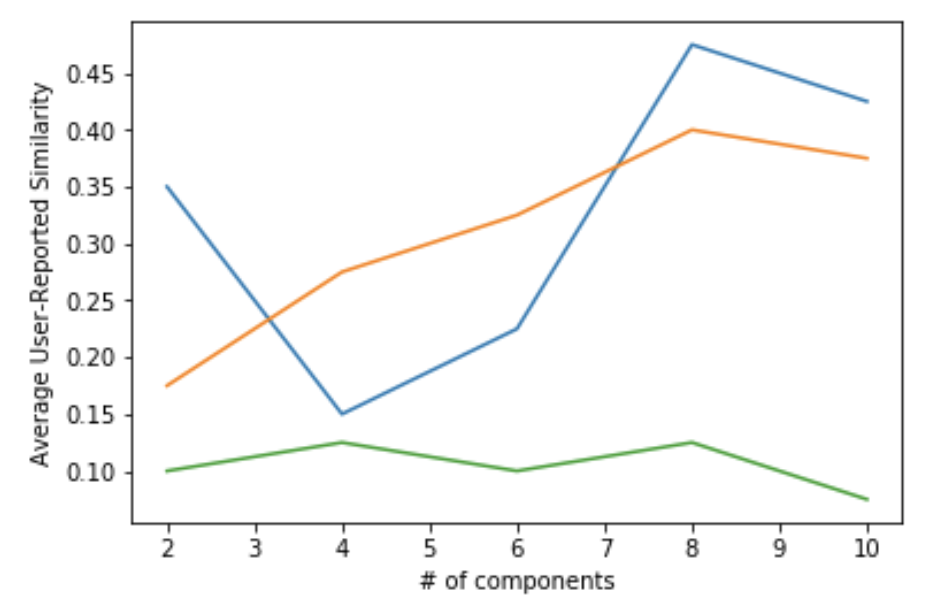}
    \caption{Average user-reported similarity scores (n=4) measured over randomly selected pairs of tweets from the new embedding space that the model marked as highly similar, versus the number of components that are selected. Blue is all appended features, orange is condensed time, and green is the original embedding space with PCA applied.}
    \label{bigboitimeboi}
\end{figure}

By comparing against manually labeled similarity scores, we found that reducing the original embedding space to 8 components prior to appending all features produced the most realistic semantic similarity metric, as displayed in Figure \ref{bigboitimeboi}. At 8 components, our model performed approximately 280\% more effectively in representing true semantic similarity than the baseline model without the addition of multimodal features. 

What could explain such a large improvement? We saw that politicians often tweet about similar topics, such as policy topics and sporting events, at similar times. Our dataset also included tweets during natural disasters, which led to many geospatially and semantically similar tweets. Furthermore, it is also worth keeping in mind that labeled data was collected in limited quantities. While it would be worthwhile to replicate this study with more labeled data, our results provide compelling evidence for the incorporation of temporal and geospatial information in analyzing tweet similarity.


\subsubsection{Qualitative Visual Analysis Using t-SNE}

\begin{figure}
    \centering
    \includegraphics[scale=0.11]{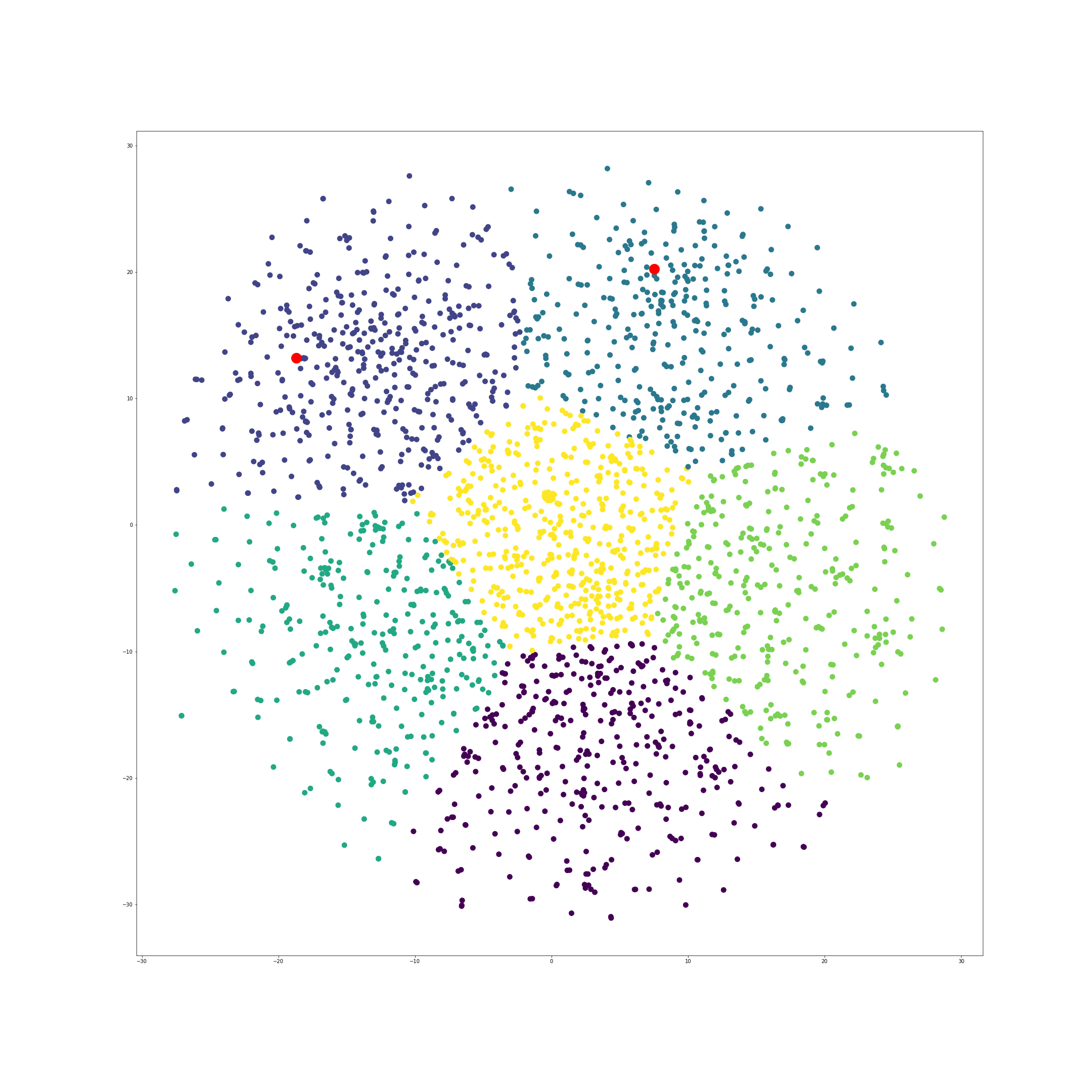}
    \caption{t-SNE applied to the original embedding space, with the two tweets marked in red.}
    \label{tsne1}
\end{figure}

\begin{figure}
    \centering
    \includegraphics[scale=0.11]{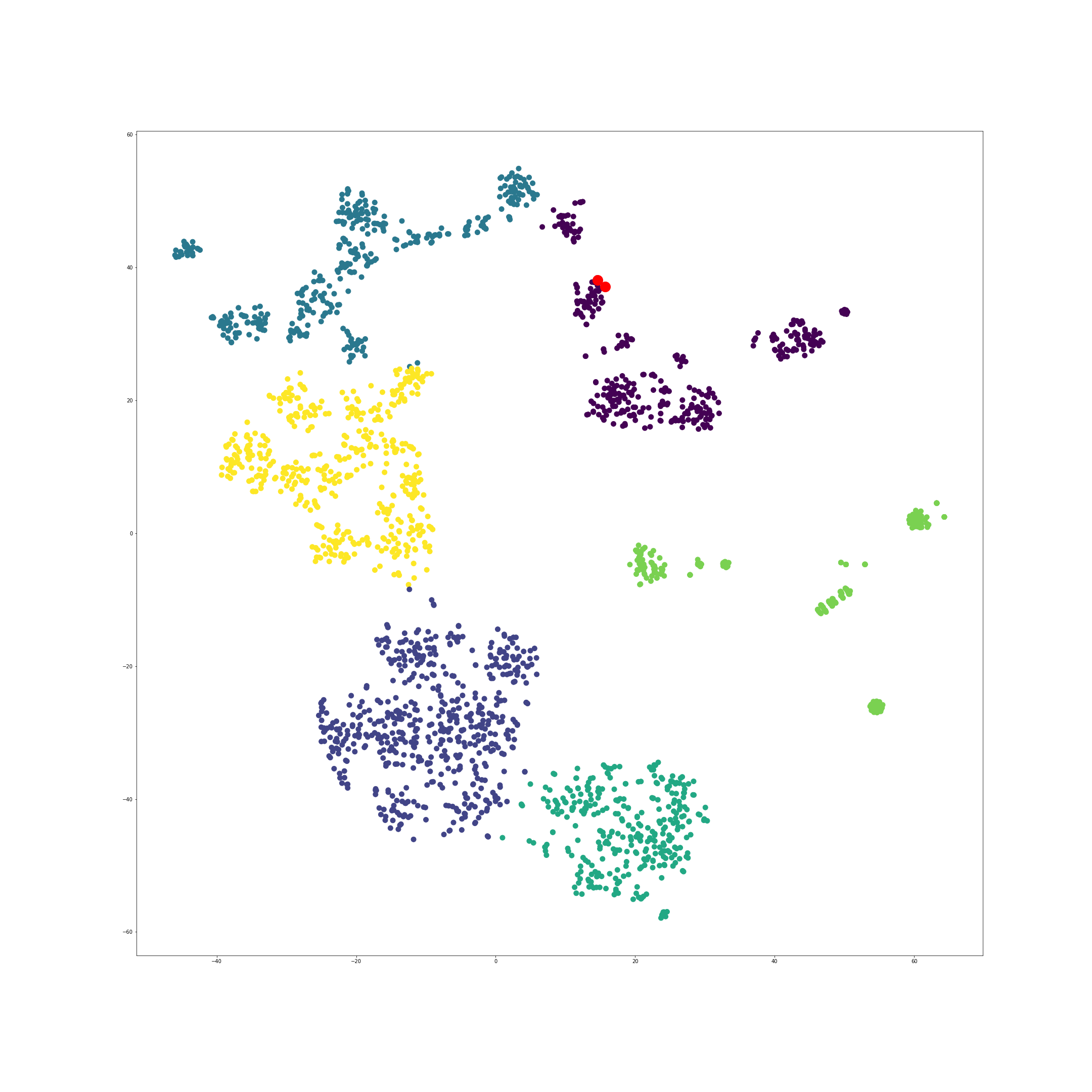}
    \caption{t-SNE applied to the modified embedding space, with the two tweets marked in red.}
    \label{tsne2}
\end{figure}

We applied t-SNE to compare the two embedding spaces visually. Figure \ref{tsne1} shows the original embedding space, containing only the embedded text of the tweets themselves. Figure \ref{tsne2} shows the embedding space after applying PCA to the embedding space and then appending all multimodal features. In both plots, the red dots represent the following two tweets:

\begin{center}
    \includegraphics[scale=0.3]{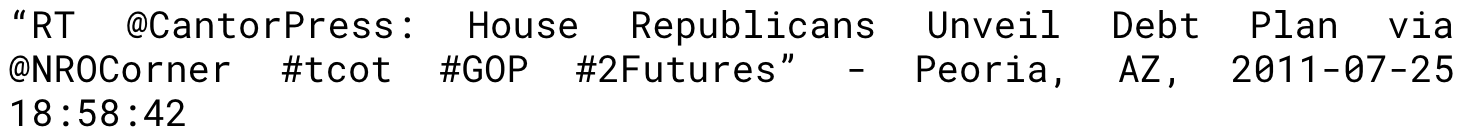}
    \includegraphics[scale=0.3]{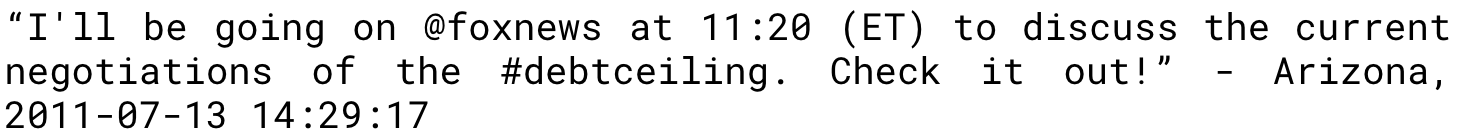}
\end{center}

These tweets are semantically similar: Both concern the U.S. debt-ceiling crisis of 2011, and their authors share similar desired policy outcomes. However, they are not textually similar, and hence are classified as dissimilar according to the distributional hypothesis taken by Doc2Vec. On the other hand, they are separated by small amounts of time and distance, and hence are significantly closer in this new space.

\begin{figure}
    \centering
    \includegraphics[scale=0.11]{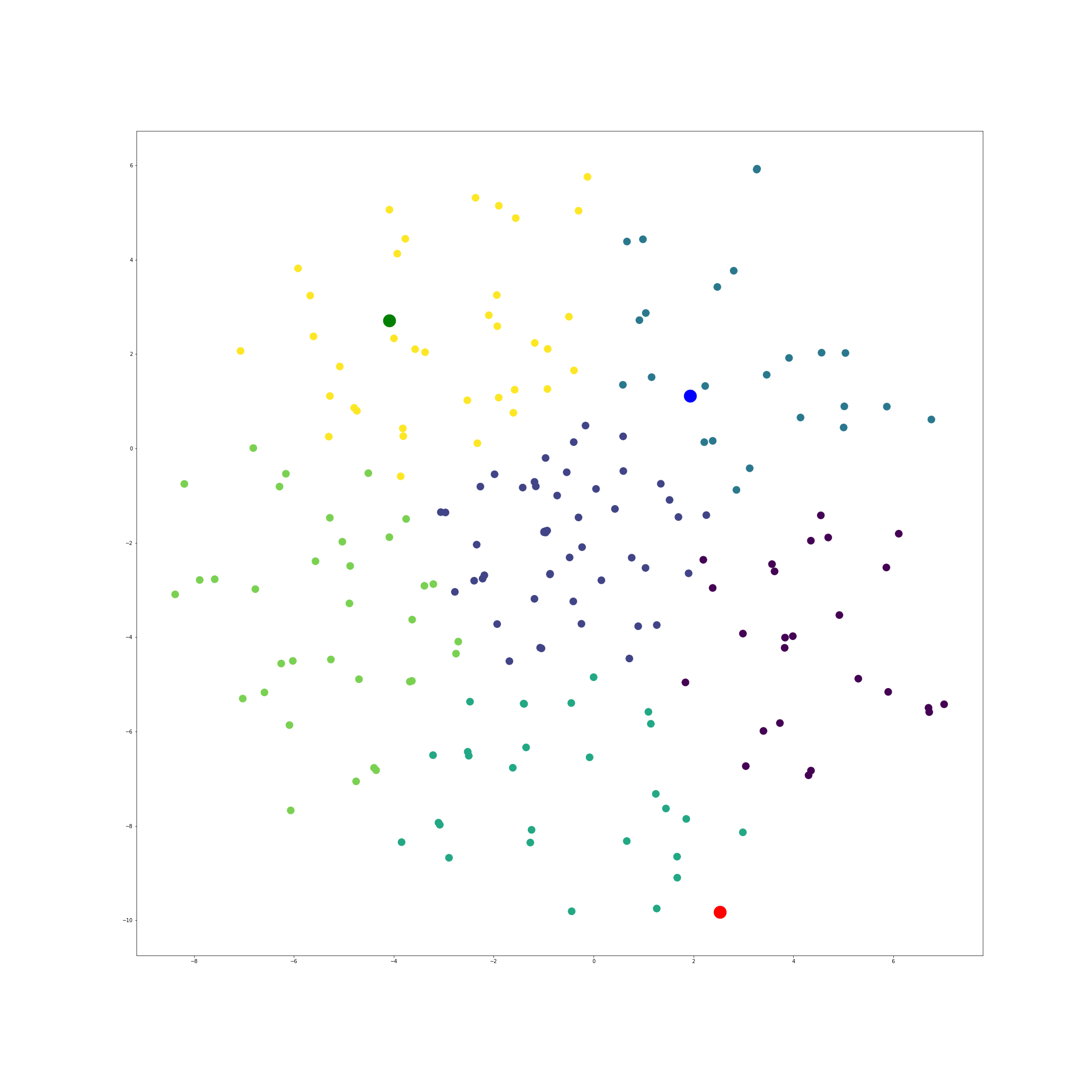}
    \caption{t-SNE applied to the original embedding space of the 2016 election subset, with the three tweets colored.}
    \label{tsne3}
\end{figure}

\begin{figure}
    \centering
    \includegraphics[scale=0.11]{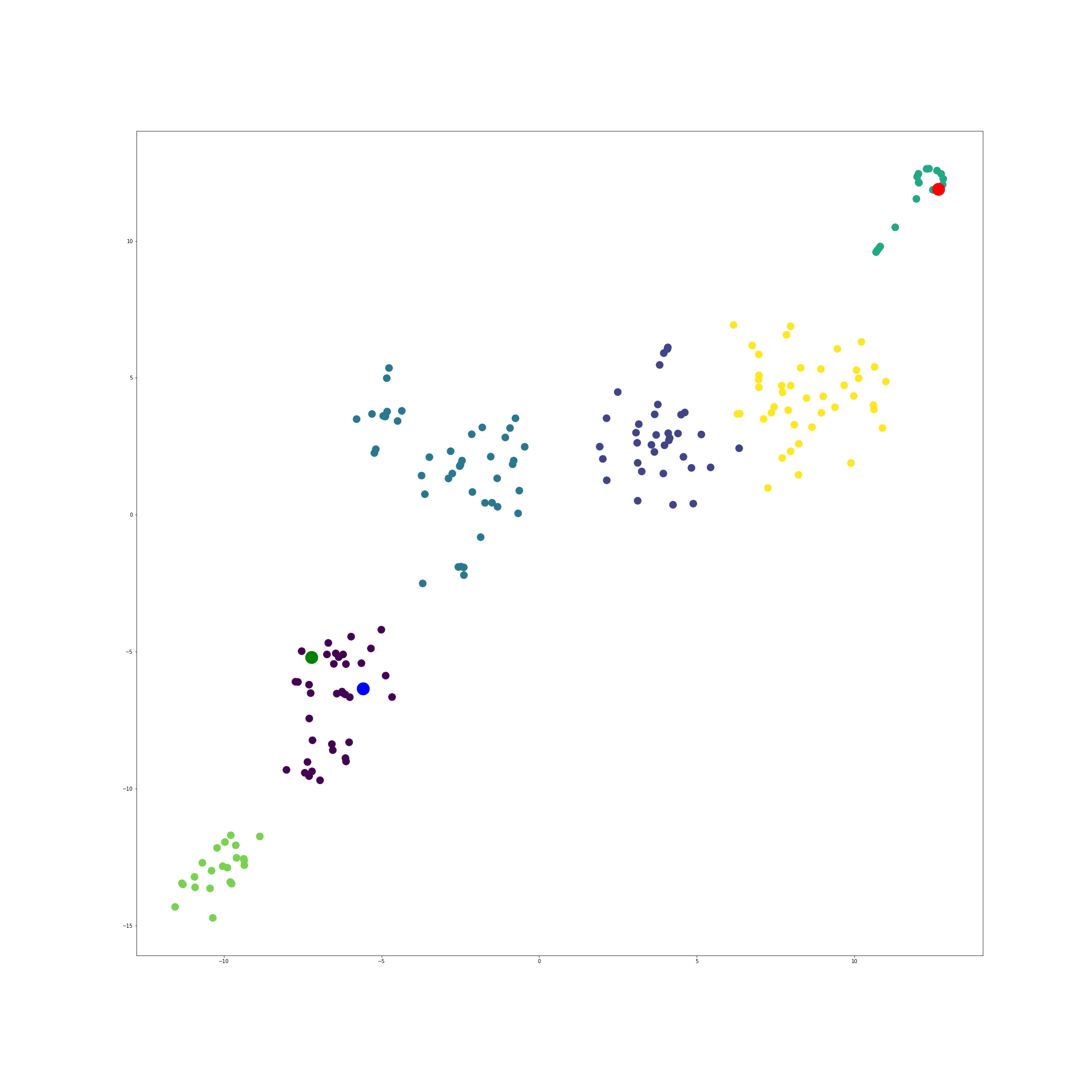}
    \caption{t-SNE applied to the modified embedding space of the subset, with the three tweets colored.}
    \label{tsne4}
\end{figure}

We also tested our method on a small subset of tweets spanning the months prior to the 2016 election. Figure \ref{tsne3} shows the original embedding space, while Figure \ref{tsne4} shows the modified space with appended features. The tweets are colored as follows:

\begin{center}
    \includegraphics[scale=0.3]{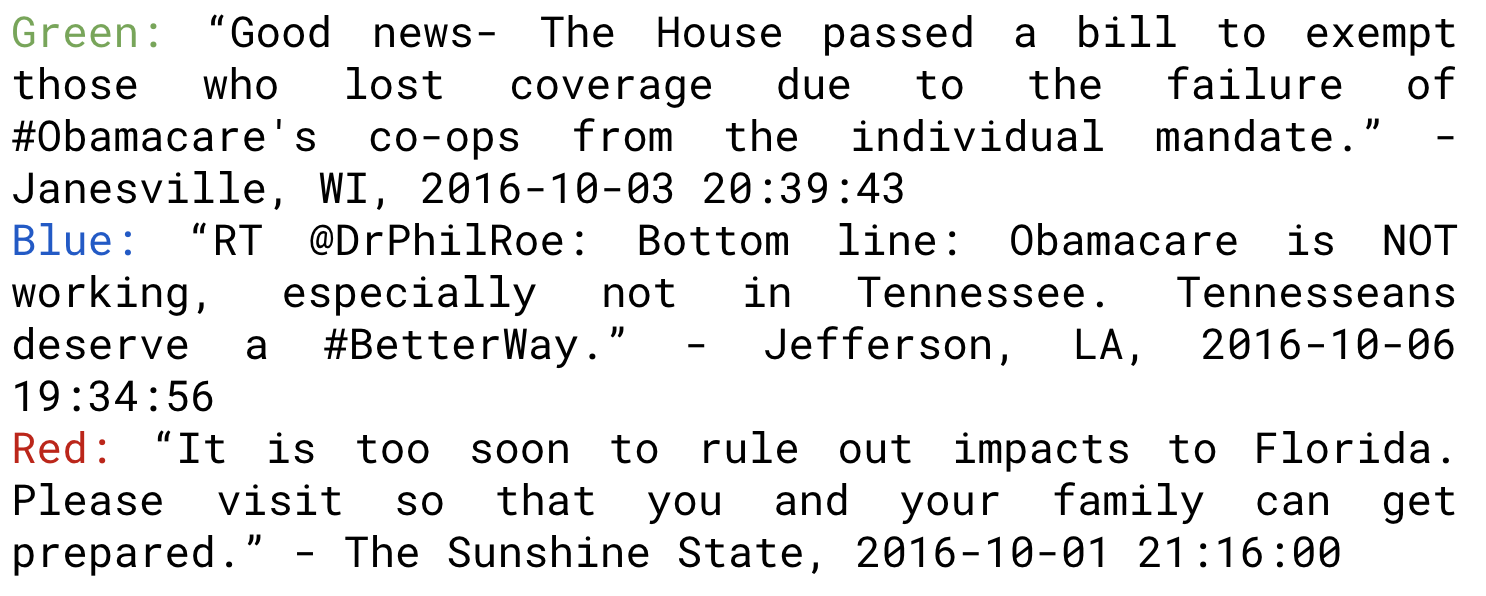}
\end{center}

Semantically, the blue tweet is much closer to the green tweet than the red tweet. In the original space, they are evenly-spaced; however, in the new space, the blue and green tweets are much closer to one another, while the red tweet has remained distant, suggesting that negative semantic relationships are also preserved under this transformation.

\section{Conclusion}
\subsection{Final Thoughts}
In this paper, we introduced methods to incorporate additional features into traditional semantic similarity algorithms. We found that reducing the dimension of the original embedding space and then appending additional non-textual features performed better than the original embedding space itself. This also performed better than iterative minimization, which we believe is due to the PCA-based approach working in more dimensions. By acting on the level of the embedding space, it incorporated multimodal attributes directly into the orientation of the tweet vectors. On the other hand, iterative minimization lacked spatial context and only acted on the final computed similarity score.

Overall, the success of our PCA-based model supports the hypothesis that multimodal data can provide valuable context for determining semantic similarity.

\subsection{Future Work}
We would like to broaden our experimentation in collecting more labeled data. We would also like to apply our algorithm in testing if tweets from local politicians differ from national politicians when controlling for location. More broadly, we would like to extend our results beyond the scope of political microblogging and apply it to other multimodal datasets. In practice, multimodal attributes are extraordinarily powerful and underutilized contextual markers, and so may prove to be quite valuable in building NLP engines of the future. 

\end{document}